\documentclass{article}

\usepackage{fbstyle}
\usepackage{graphicx}
\usepackage[margin=3cm]{geometry}
\usepackage{tikz}
\usepackage{pgfplots}
\usepackage{rotating}

\usepackage{hyperref}
\usepackage{pdfpages}
\usepackage{booktabs}

\title{Efficient Multiscale Object-based Superpixel Framework}
\author{%
    Felipe Bel{\'e}m\textsuperscript{1,2,}\footnote{felipe.belem@\{ic.unicamp.br,esiee.fr\}}, %
    Benjamin Perret\textsuperscript{2}, %
    Jean Cousty\textsuperscript{2}\\%
    Silvio J. F. Guimar{\~a}es \textsuperscript{3}, %
    Alexandre Falc{\~a}o\textsuperscript{1}%
  \vspace{0.2cm}\\
  \textsuperscript{1}~University of Campinas, Campinas, Brazil\\
  \textsuperscript{2}~Universit{\'e} Gustave Eiffel, Noisy-le-Grand, France\\
  \textsuperscript{3}~Pontifical Catholic University of Minas Gerais, Belo Horizonte, Brazil\\
}
\date{}

\begin{document}
    \maketitle

    \begin{abstract}
	  Superpixel segmentation can be used as an intermediary step in many applications, often to improve object delineation and reduce computer workload. However, classical methods do not incorporate information about the desired object. Deep-learning-based approaches consider object information, but their delineation performance depends on data annotation. Additionally, the computational time of object-based methods is usually much higher than desired. In this work, we propose a novel superpixel framework, named \textit{Superpixels through Iterative CLEarcutting} (SICLE), which exploits object information being able to generate a multiscale segmentation on-the-fly. SICLE starts off from seed oversampling and repeats optimal connectivity-based superpixel delineation and object-based seed removal until a desired number of superpixels is reached. It generalizes recent superpixel methods, surpassing them and other state-of-the-art approaches in efficiency and effectiveness according to multiple delineation metrics.
    \end{abstract}

  \section{Introduction}
\label{sec:introduction}

	Some algorithms partition an image into several disjoint groups of connected pixels, named \textit{superpixels}. The pixels for the superpixels share a common property (\fbeg, color or texture). Thus, a superpixel-based image representation provides more contextual information than a pixel-based one while reducing the computer workload. Due to these properties, superpixel segmentation is often used as an intermediary step in medical applications~\cite{Dhore:2021:Xray,Zhou:2019:Breast,Liu:2019:LungSegmentation}, semantic segmentation~\cite{Yi:2022:Semantic}, plant diseased-leaf detection~\cite{Zhang:2018:Plant}, pedestrian segmentation~\cite{Yu:2019:Pedestrian}, among other examples.

	Most methods model the superpixel segmentation problem as a pixel-clustering task. The most popular one is \textit{Simple Linear Iterative Clustering} (SLIC)~\cite{Achanta:2012:SLIC} whose efficiency and simplicity inspired several and more accurate algorithms~\cite{Liu:2018:IMSLIC,Li:2015:LSC}. Other methods, like \textit{Superpixel Hierarchy} (SH)~\cite{Wei:2018:SH}, \textit{Entropy Rate Superpixels} (ERS)~\cite{Liu:2011:ERS}, initial approaches based on \textit{Iterative Spanning Forest} (ISF) framework~\cite{Vargas:2019:ISF}, and \textit{Dynamic ISF} (DISF)~\cite{Belem:2020:DISF}, model the problem differently and offer a more effective solution with slightly higher computational cost. However, all the aforementioned methods are unaware of the object of interest in the image. Thus, instead of ensuring the boundaries of interest (\fbie, relevant borders), they pursue the preservation of all detectable ones, including many irrelevant ones.
	
	Recently, several deep-learning-based approaches have been proposed for generating superpixels in which the breakthrough of end-to-end trainable methods was \textit{Superpixel Sampling Networks} (SSN)~\cite{Jampani:2018:SSN}, further improved in~\cite{Yang:2020:FCN}. These methods incorporate object information during training, but they require data annotation, which may be scarce in some applications (\fbeg, medical and biological). Although unsupervised methods have also been proposed~\cite{Yu:2021:EASS,Suzuki:2020:RIM}, they all present moderate boundary delineation considering different datasets, even those of the same domain (\fbeg, natural image datasets).
	
	Some algorithms consider the object information represented as a monochromatic image (\fbie, an object saliency map) in which pixel intensity denotes an object membership value. These object-based methods use such information for favoring the delineation of the object of interest at the expense of others, resulting in a more accurate superpixel segmentation. However, exhaustively exploiting saliency information can lead to higher computational cost and saliency quality dependency, as seen in \textit{Object-based Iterative Spanning Forest} (OISF)~\cite{Belem:2018:OISF,Belem:2020:OISF} variants. Conversely, \textit{Object-based DISF} (ODISF)~\cite{Belem:2021:ODISF} overcomes the latter drawback while still being slightly slower than object-unaware methods.
	
    Due to SLIC's simplicity and efficiency, the vast majority of superpixel methods adopt the same three-step pipeline: (i) seed sampling; (ii) superpixel generation; and (iii) seed recomputation; in which step (i) is applied once and, subsequently, steps (ii) and (iii) are repeated for a given number of iterations. The major advantage of seed-based superpixel segmentation is being applicable for any object, exploiting its inherent existence in the image. Moreover, by recomputing the seeds (for the next iteration), the intra-superpixel dissimilarity and, thus, errors are minimized. Despite SLIC presents fair effectiveness, recent and more accurate approaches benefit from variations of the same pipeline~\cite{Li:2015:LSC,Liu:2018:IMSLIC,Vargas:2019:ISF,Xiao:2018:CAS}. 
   
   	This three-step pipeline presents several drawbacks. First, the number of seeds in step (i) approximates the desired number of superpixels. While it may not choose seeds that promote accurate object delineation (\fbie, relevant), especially when the object is small, using object-based strategies usually lead to higher computational time~\cite{Belem:2019:OSMOX}. Second, several executions of the above pipeline are required to produce a multiscale superpixel segmentation~\cite{Wei:2018:SH}. Third, reallocating a seed may not guarantee its accurate delineation in a previous iteration. Finally, imposing a strict number of iterations may force the execution of unnecessary iterations, notably when no convergence is met.

    A novel strategy based on the previous pipeline was recently proposed, exploited by DISF and ODISF. In step (i), a high initial number of seeds is selected for increasing the chances of sampling relevant seeds through faster approaches. A portion of the most irrelevant seeds is removed at each iteration in step (iii), based on an exponential curve, until reaching the desired number of superpixels. Thus, the most relevant ones are rest unaltered for accurate delineation in the next iteration. As one may notice, this new approach of oversampling and iterative seed removal, by design, permits obtaining a multiscale segmentation in a single execution. Moreover, as the desired number of superpixels increases, fewer iterations are performed for segmentation. However, although a strict quantity of iterations is not imposed, such number is indirectly defined by the initial seed quantity, making it challenging to be intuitively set.

\begin{figure}[ht!]
		\centering
		\includegraphics[width = \textwidth]{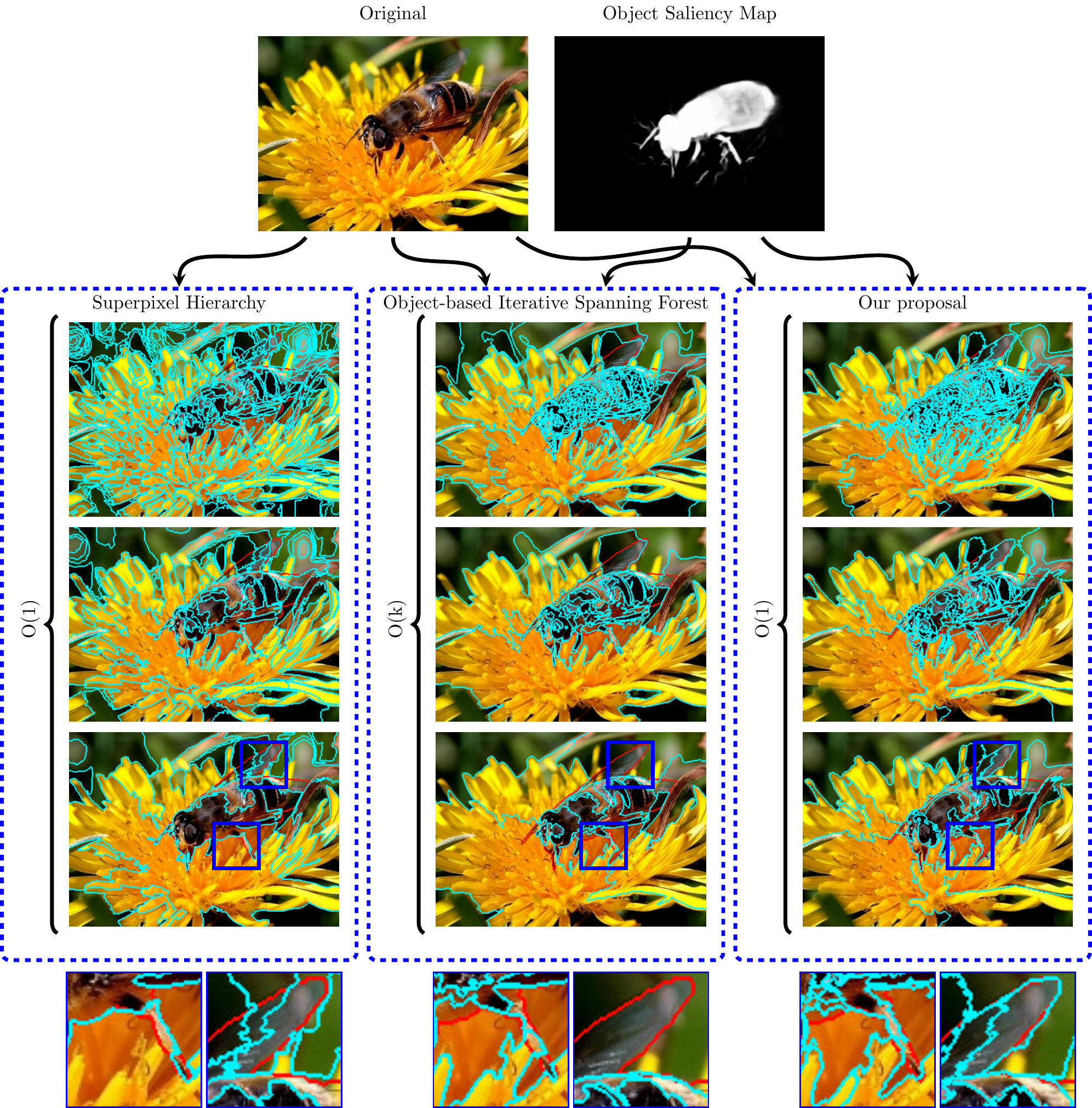}
		\caption{Multiscale superpixel segmentation comparison between two state-of-the-art methods and our proposal. It was required three scales (\fbie, $k=3$): 500, 100, and 25 superpixels, respectively. The red lines indicate the object boundary, whereas the cyan ones, superpixel borders.}
		\label{fig:intro}
	\end{figure}

	In this work, we propose a \textbf{generalization} of DISF and ODISF, overcoming their inability to directly control the number of iterations for segmentation. By reformulating the seed removal step, our proposal, named \textit{Superpixels through Iterative CLEarcutting} (SICLE), is capable of limiting the algorithm to perform a maximum number of iterations, such that it can obtain DISF's and ODISF's equivalent results if the seed removal decay is appropriately set. Moreover, by not incorporating saliency information during delineation, SICLE is robust to incorrect saliency estimations while producing highly accurate superpixels. Finally, SICLE can generate an object-based multiscale segmentation on-the-fly if the quantity of seeds at each iteration is previously defined  (see Figure~\ref{fig:intro}). We also extensively evaluate different configurations (including new proposals) for each pipeline's step. Results show that SICLE can accurately delineate boundaries of various objects from distinct datasets, being faster than SLIC in most practical cases.

	Our main contributions may be summarized as follows:
	\begin{description}
		\item[\textbf{Seed sampling}] An equidistant {seed sampling} strategy that distributes seeds proportionally along each image axis;
		\item[\textbf{Seed removal}] Regarding seed removal, we proposed a (i) new criteria that accurately maintains those promoting effective object delineation; and (ii) a  procedure which establishes a maximum non-strict number of iterations for improving speed by avoiding unnecessary iterations for segmentation;
		\item [\textbf{Multiscale approach}]A novel framework (SICLE) for the design of object-based multiscale superpixel segmentation methods;
		\item [\textbf{On-the-fly}]SICLE-based methods that are robust to object saliency errors and dismiss unnecessary segmentation iterations while generating multiscale boundaries on-the-fly;
		\item[\textbf{Easy-to-use}] Optimization and extensive analysis of each SICLE's step, leading to an easy-to-use framework with accurate and fast delineation in different image domains;
	\end{description}

	This paper is organized as follows. In Section~\ref{sec:related}, we present an overview of the state-of-the-art in superpixel segmentation. We describe our theoretical background in Section~\ref{sec:theobkg} and, subsequently, detail our proposed framework in Section~\ref{sec:method}. Finally, in Sections~\ref{sec:results} and~\ref{sec:conclusion}, we discuss the experimental results and draw conclusions and possible future work, respectively.

  \section{Related Works}
\label{sec:related}

	In this section, we present some of the most recent and notable works in superpixel segmentation. One may refer to~\cite{Stutz:2018:Superpixels} for a deeper discussion on this topic. In Section~\ref{sec:related:classic}, we review those methods that do not incorporate any prior object information. Subsequently, we present those which use deep-learning strategies for generating superpixels in Section~\ref{sec:related:deep}. Lastly, in Section~\ref{sec:related:objbased}, we discuss the algorithms that consider object information represented by an object saliency map. Table~\ref{tab:related} compares different state-of-the-art methods against our proposal concerning some properties.

	\begin{table}[htbp!]
		\centering
		{%
\setlength{\tabcolsep}{2.5pt}
\resizebox{\linewidth}{!}{
\begin{tabular}{c||c|c|c|c|c|c}\toprule
  Properties & SLIC\cite{Achanta:2012:SLIC} & LSC\cite{Li:2015:LSC} & SH\cite{Wei:2018:SH} & SEAL\cite{Tu:2018:SEAL}\textsuperscript{1} & OISF\cite{Belem:2020:OISF} & SICLE \\\midrule\midrule
  Boundary Recall & 0.864 & 0.893 & 0.949 & 0.826 & 0.952 & 0.978 \\\hline
  Under-segmentation Error & 0.016 & 0.013 & 0.011 & 0.015 & 0.008 & 0.010 \\\hline
  Speed~(ms) & 507 & 850 & 941 & -\textsuperscript{2} & 2832 & 468 \\\hline
  Multiscale Segmentation & No & No & Yes & No & No & Yes \\\hline
  Object Information & No & No & No & Yes & Yes & Yes \\\hline
  Error Propagation & No & No & Yes & Yes & Yes & No \\\bottomrule
\end{tabular}
}
\begingroup
\renewcommand{\arraystretch}{0.5}
\scriptsize\raggedleft
\begin{tabular}{p{0.95\textwidth}}
  \textsuperscript{1}~Deep-learning-based algorithm.\\
  \textsuperscript{2}~\mbox{For a single segmentation, the time for loading the network exceeds OISF's.}
\end{tabular}
\endgroup
}
		\caption{Average SICLE performance compared to state-of-the-art superpixel algorithms for 750 superpixels. It was considered a grayscale dataset whose images are 512$\times$512.}
		\label{tab:related}
	\end{table}

	\subsection{Classical Methods}
	\label{sec:related:classic}

		We may broadly classify the majority of the classical methods by their strategy: (i) clustering-based; (ii) graph-based; and (iii) path-based. In the first group, the superpixels are generated through pixel clustering. \textit{Simple Linear Iterative Clustering} (SLIC)~\cite{Achanta:2012:SLIC} is the most popular method of such group, and it uses an adapted K-means on a 5-dimensional feature space for a fast but moderate delineation. Similarly, \textit{Linear Spectral Clustering} (LSC)~\cite{Li:2015:LSC} applies a weighted K-means in a deliberately tailored 10-dimensional space for producing superpixels with high boundary adherence while slightly increasing the computational time. Some works adopt different clustering techniques for superpixel generation, such as Gaussian Mixture Models~\cite{Ban:2018:GMM} or Density-based Spatial Clustering of Applications with Noise~\cite{Shen:2016:DBSCAN}. A common drawback of methods within such group is not ensuring the desired number of superpixels.

		We may list \textit{Entropy Rate Superpixels} (ERS)~\cite{Liu:2011:ERS} and the \textit{Superpixel Hierarchy} (SH)~\cite{Wei:2018:SH} as the most popular and the most effective graph-based methods, respectively. In ERS, the superpixel segmentation is modeled as an optimization problem of the entropy of a random walk in the graph's topology. While ERS presents high computational time, even when using greedy strategies, SH uses the Bor{\r u}vka algorithm for generating superpixels with high boundary adherence in linear time. Given it is a hierarchical method, errors in the lower levels are propagated to the upper ones, compromising its delineation performance.

		Path-based approaches define superpixels by path searching from each seed to its most similar pixels, ensuring superpixel connectivity. The methods based on the \textit{Image Foresting Transform} (IFT)~\cite{Falcao:2004:IFT} algorithm often present top object delineation at the expense of not easily controlling the superpixel compacity~\cite{Stutz:2018:Superpixels}. \textit{Iterative Spanning Forest} (ISF)~\cite{Vargas:2019:ISF} is a three-step seed-based framework that generates superpixels through iterations of IFT executions on improved seed sets. Based on the latter, the authors in~\cite{Galvao:2018:RISF} proposed the \textit{Recursive ISF} (RISF) algorithm to obtain a sparse-hierarchical superpixel segmentation with no efficiency deterioration. Recently, the authors in~\cite{Belem:2020:DISF} proposed \textit{Dynamic ISF} (DISF), which starts from seed oversampling and iteratively removes seeds based on a relevance criterion applied to the IFT-based superpixel delineations, achieving highly accurate segmentation. Although this group also contains non-IFT-based methods~\cite{Achanta:2017:SNIC}, they usually present fair delineation performance.

		Some methods aim to increase the number of superpixels in regions characterized by higher content density (\fbie, \textit{content-sensitive}). For instance, \textit{Intrinsic Manifold SLIC} (IMSLIC)~\cite{Liu:2018:IMSLIC} maps every pixel to a 2-dimensional manifold and measures the superpixel density through its area. Other examples, like \textit{Content-Adaptive Superpixels} (CAS)~\cite{Xiao:2018:CAS} and \textit{Texture-Aware and Structure-Preserving} (TASP)~\cite{Wu:2021:TASP}, generate content-sensitive superpixels by improving the separability in the feature space. However, differently from IMSLIC, CAS and TASP require a post-processing step for ensuring superpixel connectivity.

	\subsection{Deep-learning Methods}
	\label{sec:related:deep}

		Although classical methods often present high delineation accuracy, they are unaware of the user's desire to segment a particular object. On the other hand, deep-learning methods are driven by the information inferred during training (\fbeg, important borders). For a period of time, some works~\cite{Awaisu:2019:DeepFLIC} aimed at generating superpixels using deep features, even though the benefits of this strategy are not yet solid~\cite{Tu:2018:SEAL}. Thus, in~\cite{Tu:2018:SEAL}, the authors proposed a novel loss function named \textit{SEgmentation Aware Loss} (SEAL) for learning the pixel affinities for the underlying ERS algorithm. Such method, named SEAL-ERS, was further improved in~\cite{Peng:2021:HERS} in terms of speed. Still, as one may argue, SEAL-ERS is dependable on the quality of estimating such affinities since incorrect inferences can prejudice the entropy computation performed by its backbone method.

		Other common approach is to design end-to-end trainable networks for superpixel segmentation, either supervisedly or unsupervisedly. Although both are dependable on the quantity of data available for training, the latter does not require any data annotation. We may cite the popular \textit{Superpixel Sampling Networks} (SSN)~\cite{Jampani:2018:SSN} and \textit{Fully Convolutional Network} (FCN)~\cite{Yang:2020:FCN} as examples of supervised methods. Regarding unsupervised ones, we list \textit{Regular Information Maximization} (RIM)~\cite{Suzuki:2020:RIM} and \textit{Edge-Aware Superpixel Segmentation} (EASS)~\cite{Yu:2021:EASS} as two of the most recent representatives of this group.

	\subsection{Object-based Methods}
	\label{sec:related:objbased}

		We term an algorithm as object-based when it incorporates prior object information, often represented by object saliency maps. \textit{Saliency-based Superpixels} (SS)~\cite{Xu:2014:SS} can be classified as such, even though such term was first coined in~\cite{Belem:2018:OISF}. Recent object-based solutions can be seen as generalizations of previous methods for handling with saliency maps, such as \textit{Object-based ISF} (OISF)~\cite{Belem:2018:OISF,Belem:2020:OISF}. In OISF, the user controls the superpixel displacement and morphology concerning the map's estimation. While it is slow and it is highly dependable on the map quality (\fbie, propagates the maps' errors), \textit{Object-based DISF} (ODISF)~\cite{Belem:2021:ODISF} offers a highly accurate and faster solution with minimum influence of saliency errors.

  \section{Theoretical Background}
\label{sec:theobkg}
  In this section, we present important definitions regarding our proposal. First, we recall basic graph notions in Section~\ref{sec:theobkg:imggraph} and, subsequently in Section~\ref{sec:theobkg:ift}, we detail the \textit{Image Foresting Transform}~(IFT)~\cite{Falcao:2004:IFT}, the core algorithm for delineation.

  \subsection{Image Graph}
  \label{sec:theobkg:imggraph}

    An \textit{image} $I$ can be defined as a pair $\fbiangl{\PixelSet,\FeatFunc}$ in which $\PixelSet \subset \fbzset^2$ is the set of \textit{picture elements}~(\fbie, pixels) and $\FeatFunc$ maps every $p \in \PixelSet$ to its \textit{feature vector} $\FeatFunc(p) \in \fbrset^m$, given $m \in \fbnset^*$. If $m>1$, $I$ is a \textit{colored} image (\fbeg, CIELAB), otherwise, $I$ is \textit{grayscale}. Lastly, an \textit{object saliency map} $O = \fbiangl{\PixelSet,\ObjFunc}$ is a grayscale image whose feature indicates a proportional likelihood $\ObjFunc(p) \in \fbisqbrk{0,1}$ of a pixel $p$ belonging to an object of interest. 

    An \textit{image digraph} $G$ derived from $I$ is a pair $\fbiangl{\VertexSet,\ArcSet}$ such that $\VertexSet \subseteq \PixelSet$ is the \textit{vertex} set and $\ArcSet \subset \VertexSet^2$ is the set of \textit{arcs}. Two vertices $x,y$ are said to be \textit{adjacents} if $\exists \fbiangl{x,y}$ or $\fbiangl{y,x} \in \ArcSet$. A common approach for establishing the arcs in $\ArcSet$ is based on the Euclidean spatial distance between the vertices. Therefore, for any $x,y \in \VertexSet$ and a radius $r \in \fbrset_+$, we may obtain the arc set $\ArcSet^r = \fbibrk{\fbiangl{x,y} : \fblnorm{2}{x - y} \leq r}$. As one may see, $\ArcSet^{\sqrt{2}}$ contains the arcs defined in an 8-neighborhood. Throughout this paper, $G$ has neither loops nor parallel edges (\fbie, $G$ is a \textit{simple digraph}).

    We may define a \textit{(directed) path} $\rho$ as a sequence of distinct vertices $\fbiangl{v_i}_{i=1}^{k}$ in which $\fbiangl{v_i,v_{i+1}} \in \ArcSet$ for $i < k$. $\rho$ is \textit{trivial} when $k=1$, and \textit{non-trivial}, otherwise. For $1 < i < k$, $v_i$ is said to be the \textit{predecessor} of $v_{i+1}$ and the \textit{sucessor} of $v_{i-1}$. We may refer to $v_1$ and $v_k$ as \textit{origin} (or \textit{root}) and \textit{terminal} vertices of $\rho$, respectively. For simplicity, we can exhibit both either by $\rho_{v_1 \rightsquigarrow v_k}$ or by $\rho_{v_k}$. Finally, $\rho_y = \rho_x \odot \fbiangl{x,y}$ denotes the path $\rho_y$ resultant from concatenating $\rho_x$ with an arc $\fbiangl{x,y}$.

  \subsection{Image Foresting Trasform}
  \label{sec:theobkg:ift}

    The \textit{Image Foresting Transform}~(IFT)~\cite{Falcao:2004:IFT} is an elegant framework for developing connectivity-based image operators, and can be computed using a generalization of the Dijkstra's algorithm. Due to its object delineation performance, it has been used in several applications~\cite{Martins:2019:SymmISF,Sousa:2019:ALTIS,Castelo:2019:ClassISF}, especially for superpixel segmentation~\cite{Vargas:2019:ISF,Belem:2020:DISF,Galvao:2018:RISF}. In this work, we consider its seed-restricted variant. That is, for a given \textit{seed set} $\SeedSet \subset \VertexSet$, the IFT find paths with optimum cost from any seed to every $p \in \VertexSet\setminus\SeedSet$.

    A \textit{path-cost function} $\ConnFunc_*$ assigns a cost $\ConnFunc_*(\rho) \in \fbrset_+$ to any $\rho \in \PathSet$, considering $\PathSet$ to be the set of all possible paths in $G$. Similarly to $\ConnFunc_*$, an \textit{arc-cost function} $\ArcFunc_*(x,y) \in \fbrset$ defines a cost to any arc $\fbiangl{x,y} \in \ArcSet$. A popular option of path-cost function is the $\ConnFunc_{\max}$ function (Equation~\ref{eq:theobkg:fmax}) due to its accurate object delineation performance~\cite{Belem:2020:DISF,Bragantini:2018:DynIFT}: 
    \begin{equation}
      \begin{aligned}
        \ConnFunc_{max} (\fbiangl{x}) &= %
          \begin{cases}
            \begin{aligned}
              0, & \quad \text{if } x \in \SeedSet\\
              +\infty, & \quad \text{otherwise}\\
            \end{aligned}
          \end{cases} \\
        \ConnFunc_{max}(\rho_x \odot \fbiangl{x,y}) &= \max\fbibrk{\ConnFunc_{max}(\rho_x), \ArcFunc_*(x,y)}
      \end{aligned}
      \label{eq:theobkg:fmax}
    \end{equation}
    A path $\rho_x$ is said to be \textit{optimum} if $\ConnFunc_*(\rho_x) \leq \ConnFunc_*(\tau_x)$ for any other $\tau_x \in \PathSet$ irrespective of its origin. 

    By electing a unique optimum-path $\rho_x$ for all $x \in \VertexSet$, we define the \textit{predecessor map} $\PredMap$ as an acyclic map which assigns $x$ either to its predecessor in $\rho_x$ or to a to a distinctive marker $\NilVertex \not \in \VertexSet$ when $x$ is the root of $\rho_x$ (and, consequently, of $\PredMap$). Note that we can recursively map $x$ to its root $\RootMap(x) \in \SeedSet$ through $\PredMap$. Starting at the seeds, the IFT framework builds $\PredMap$ through path concatenation while it minimizes a \textit{cost map} $\CostMap(x) = \min_{\rho_x \in \PathSet}\fbibrk{\ConnFunc_*(\rho_x)}$. Even if $\ConnFunc_*$ does not satisfy certain properties, the resulting paths still present desirable properties for superpixel segmentation~\cite{Mansilla:2016:BirdsInsects}. For any $s \in \SeedSet$, $\PredMap$ defines an \textit{optimum-path tree} (or, equivalently, a superpixel) $\Tree(s) = \fbibrk{x : \RootMap(x) = s}$ rooted in $s$ whose paths are more closely connected to its seed than to any other. Finally, we can calculate its mean feature vector $\MeanFeatFunc(\Tree(s))$ and mean saliency value $\MeanObjFunc(\Tree(s))$ by $\sum_{x \in \Tree(s)}\FeatFunc(x)/|\Tree(s)|$ and $\sum_{x \in \Tree(s)}\ObjFunc(x)/|\Tree(s)|$, respectively.

  \section{SICLE Framework}
\label{sec:method}

\begin{figure}[ht!]
		\centering
		\includegraphics[width = \textwidth]{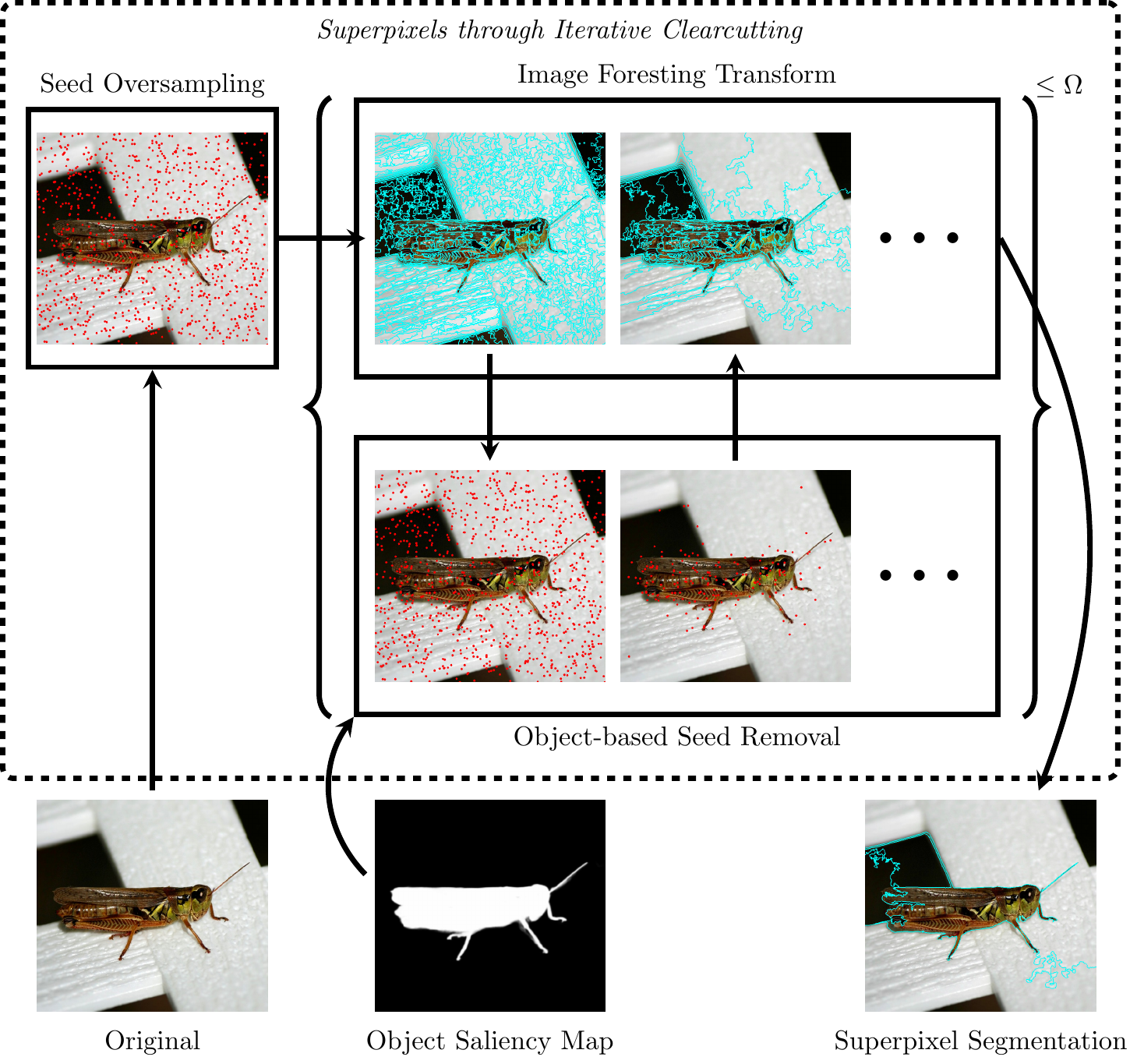}
		\caption{SICLE framework starting from $\Ninit = 1000$ and requiring $\Nfinal = 5$ superpixels.}
		\label{fig:method:pipeline}
	\end{figure}

	In this section, we describe our proposed framework named \textit{Superpixels through Iterative CLEarcutting}~(SICLE). Sections~\ref{sec:method:oversampling},~\ref{sec:method:superpixel}, and~\ref{sec:method:removal} detail, in this order, the steps of the pipeline shown in Figure~\ref{fig:method:pipeline}. 

  	\subsection{Seed Oversampling}
  	\label{sec:method:oversampling}

		The first step of SICLE is to select $\Ninit$ initial seeds for generating $\Nfinal$ superpixels, given $\Ninit,\Nfinal \in \fbnset^*$. Most seed-based algorithms aim to build a set $\SeedSet \subset \VertexSet$ containing the seeds that promote accurate object delineation (\fbie, \textit{relevant}) using the orthodox approach of sampling $\Ninit \approx \Nfinal$ seeds, making precision crucial. However, while object-based methods use a precise but slow sampling algorithm~\cite{Belem:2018:OISF,Belem:2019:OSMOX}, non-object-based ones paradoxically opt for a fast but imprecise procedure~\cite{Achanta:2012:SLIC,Liu:2018:IMSLIC,Li:2015:LSC,Vargas:2019:ISF}. Thus, although precision can lead to more effective delineation, it is inversely correlated to the algorithm's efficiency.

		On the other hand, we argue that \textit{oversampling} (\fbie, $\Ninit \gg \Nfinal$) offers a better solution since it pursues recall over precision. Given such high quantity, it is expected that a portion of the seeds in $\SeedSet$ are relevant, and, since seeds with similar features and within the same region tend to be similarly relevant, it is unnecessary to select all relevant ones. Given that, we infer that the drawbacks of using fast but imprecise sampling strategies are overcome by oversampling. Thus, the objective is to guarantee the presence of the relevant seeds while removing the irrelevant ones throughout the iterations for ensuring $\Nfinal$ superpixels in the last iteration (see Section~\ref{sec:method:removal}).

		A classic sampling strategy, hereafter named GRID~\cite{Achanta:2012:SLIC}, is to select equally-distanced seeds in a grid-like pattern. However, in most works, the image's axis are assumed to be equal, leading to an uneven distribution of the seeds. In this work, we reformulate such approach for taking into account any image dimensionality. Let $\AxisProp_* \in\; ]0,1[$ be the proportion of a given axis over all image's axis. Then, $\Napprox = (\AxisProp_X\cdot c)\cdot(\AxisProp_Y \cdot c)$ seeds are sampled considering a constant $c = \sqrt{\Ninit/\fbipar{\AxisProp_X \cdot \AxisProp_Y}}$. It is possible to determine the stride between seeds in each axis by dividing its length by $c\cdot\AxisProp_*$. As one may notice, $\Napprox \approx \Ninit$. Unlike most methods, in this work, we do not shift seeds to the lowest gradient position (in the neighborhood), given their questionable importance in oversampling.

		However, one may question the benefits of GRID in oversampling. That is, as $\Ninit$ increases, the stride between seeds decreases sharply (\fbie, they tend to be closer) and, as consequence, dispersing seeds equidistantly may not lead to improvements in delineation. Also, such strategy can become challenging if the user provides a non-rectangular mask representing a region-of-interest to be segmented: it is necessary to ensure the equidistance rule and best approximate $\Napprox$ to the initial number of seeds $\Ninit$ desired by the user. Conversely, we argue in favor of random seed oversampling (RND), given its simplicity to be implemented, especially for the latter case. Furthermore, backed up by the premise of seed relevance redundancy, one may expect that such approach can generate seed sets with accurate object delineation results.

	\subsection{Superpixel Generation}
	\label{sec:method:superpixel}

\begin{figure}[ht!]
			\includegraphics[width=\textwidth]{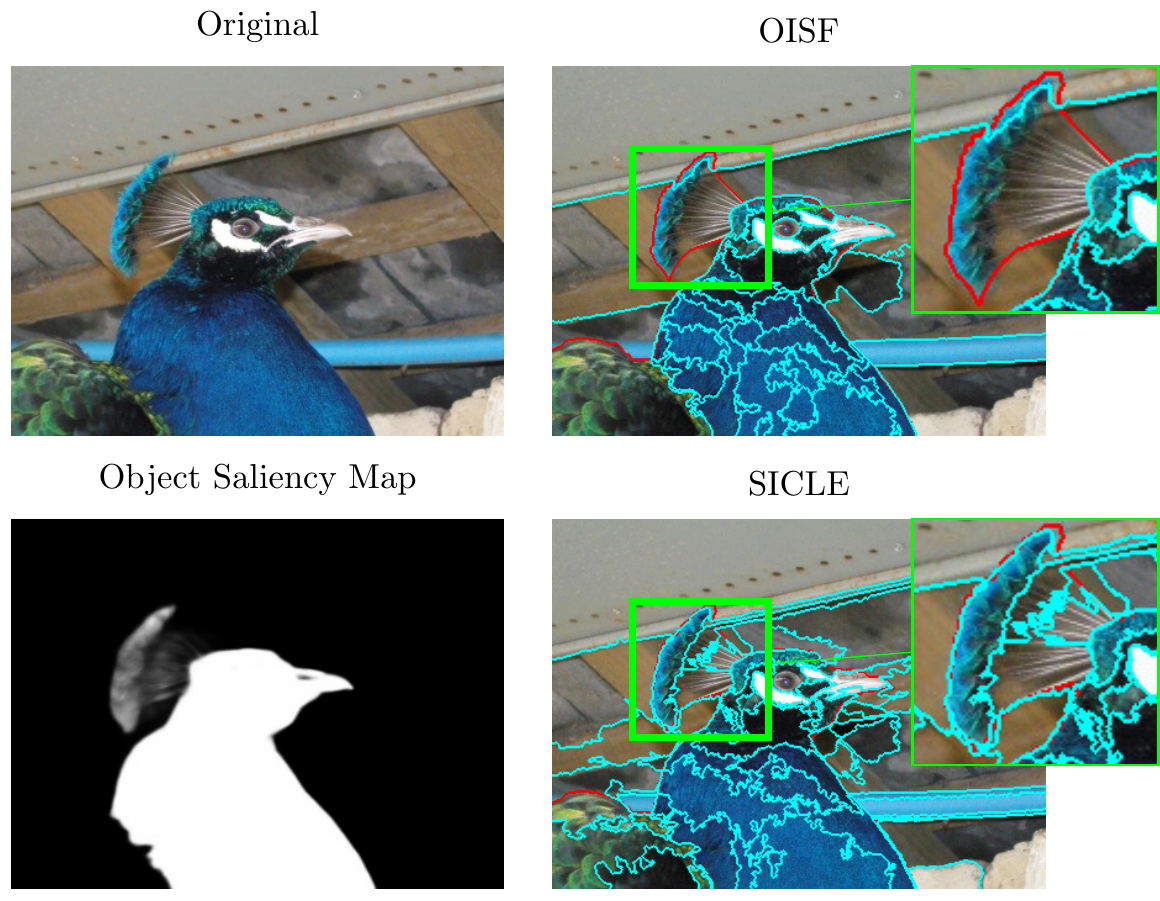}
			\caption{Comparison between SICLE and OISF regarding the dependability of the object saliency map quality. The desired number of superpixels was $\Nfinal=50$. The red lines indicate the object boundary, whereas the cyan ones, superpixel borders. }
			\label{fig:method:objsmdep}
		\end{figure}

		In SICLE, superpixels are generated by the seed-restricted version of the IFT framework. Recently, an object-based path-cost function~\cite{Belem:2018:OISF} that controls the superpixel morphology with respect to $\FeatFunc$ or $\ObjFunc$ has been proposed. However, aside from requiring parameter optimization, we claim no need to use either such function or any prior object information in delineation. Clearly, the ground-truth (\fbie, the ideal saliency map) is defined by the borders in $\FeatFunc$ during annotation by the user, and considering such map can only reinforce the preexistent borders. Also, given it is yet unfeasible to generate the ideal map, subjecting delineation to a piece of information produced by inference can deteriorate segmentation, especially by reinforcing the borders derived from incorrect estimations. Figure~\ref{fig:method:objsmdep} shows the comparison between SICLE and OISF, which uses such object-based path-cost function. Therefore, in this work, we consider the $\ConnFuncMax$ function due to its reportedly accurate performance in several works~\cite{Belem:2020:DISF,Belem:2021:ODISF,Bragantini:2018:DynIFT}.

		In~\cite{Bragantini:2018:DynIFT}, the authors propose a novel arc-cost function whose segmentation results surpassed many state-of-the-art methods in terms of interactive segmentation. The \textit{Dynamic IFT} (DYN) estimates the arc-costs with respect to the tree's mean features when it reaches the respective vertex to be conquered (\fbie, while it is \textit{growing}). More formally, the $\ArcFuncDyn$ can be defined as $\ArcFuncDyn(x,y) = \fblnorm{2}{\MeanFeatFunc(\Tree(s)) - \FeatFunc(y)}$ in which $s = \RootMap(x)$.

		However, $\ArcFuncDyn$ is highly unstable. As one may see, the tree's mean feature vector may differ for different tie-breaking policies in the IFT, which can impact the tree's composition and characteristics, and produce, thus, distinct delineations. Conversely, we argue in favor of considering the seed's features over its tree's mean vector. First, they promote stability since their features are immutable throughout the execution of the IFT. Furthermore, given that superpixels, by definition, minimize dissimilarity, one may argue that the tree's features may resemble its seed's. Therefore, both arc-cost functions may present similar delineations, especially for regions with high color dissimilarity (\fbeg, object borders). In this work, we evaluate the use of a root-based function $\ArcFuncRoot(x,y) = \fblnorm{2}{\FeatFunc(\RootMap(x)) - \FeatFunc(y)}$, hereafter named ROOT.

  \subsection{Seed Removal Criterion}
  \label{sec:method:removal}

		As presented in Section~\ref{sec:method:oversampling}, $\SeedSet$ is produced through oversampling, and it is necessary to remove $\Ninit - \Nfinal$ irrelevant seeds for generating $\Nfinal$ superpixels in the last IFT execution. Since determining whether a seed is relevant or not based on its particular features, in SICLE, we take advantage of the information generated from the previous IFT execution. That is, at each iteration, we let the seeds within $\SeedSet$ manifest their possible relevance (through competition) during the IFT and include their superpixel features in their relevance computation.

		One may note that the relevance of a seed is relatively defined (\fbeg, proximity to the object). Thus, since the ideal final set of seeds is not known beforehand, we determine the relevance of a seed by comparing it with its pairs. Then, ``perturb'' the most relevant by removing a portion of the least relevant, and test their relevance in the subsequent IFT execution. In this approach, the most relevant seeds are consistently labeled as such throughout the iterations.

		At the first iteration, it is expected that, aside from numerous irrelevant seeds, many regions contain several equally-relevant seeds. By ordering the seeds by their relevance, the portion of the least ones should present instances of the previous cases and, thus, removing a large quantity should impact the object minimally. Throughout the iterations, the number of seeds decreases, and the least relevant ones are more defined by relation (\fbie, less relevant than its pairs) than by estimation (\fbie, almost zero relevance). Moreover, fewer seeds are directly associated with reducing the seed competition, leading to more volatile superpixels and major delineation errors (especially ``leakings''). In such environment, removing smaller quantities should prevent superpixel growth instability and minimize errors. Therefore, we chose an exponential function $\SeedCurve(i) \in \fbnset^*$ for computing the quantity of relevant seeds to be maintained for the subsequent iteration $i+1 \in \fbnset^*$.

		The major drawback of this strategy is not bounding the number of iterations for reaching the desired number $\Nfinal$ of superpixels, possibly compromising its efficiency. One may argue in favor of imposing a strict number of $\Omega$ iterations for segmentation for any quantity of superpixels desired. However, we argue that the necessary number of iterations depends on $\Nfinal$. For instance, for high competition environments, intra-superpixel dissimilarity is significantly minimized and, thus, applying more iterations for subtle variations in such configuration should minimally impact the object delineation. Conversely, as $\Nfinal$ decreases, removing fewer portions of seeds (\fbie, more iterations) assists in stabilizing the superpixel growth. Therefore, we limited the number of iterations of SICLE to be at most $\Omega$ by setting $\SeedCurve(i) = \max\{(\Ninit)^{1-\omega i}, \Nfinal\}$ in which $\omega = 1/(\Omega - 1)$. For obtaining DISF and ODISF segmentations, one may set $\omega = \log_{\Ninit}e$.

		As mentioned, we exploit the optimum-path forest generated from the previous iteration for computing the relevance of each seed $s \in \SeedSet$. That is, its relevance $\SeedRel_*(s) \in \fbrset_+$ is estimated based on the characteristics of its superpixel $\Tree(s)$. Given that, in IFT, a seed's capability of conquering similar vertices is reflected on its superpixel's size, we propose a size-based criterion $\SeedRelSize(s) = |\Tree(s)|/|\VertexSet|$ for computing the relevance of $s$.

		However, since the background superpixels are often larger than the object's, $\SeedRelSize$ may undesirably favor the former. On the other hand, the typical homogeneity of the background could be differed by the heterogeneity of the object through superpixel contrast. For that, we first describe the color dissimilarity $\TreeGradFeat(s,t) \in \fbrset_+$ between two trees by $\fblnorm{2}{\MeanFeatFunc(\Tree(s)) - \MeanFeatFunc(\Tree(t))}$, given $s,t \in \SeedSet$. Then, we define the \textit{tree's neighbors} (or adjacents) by $\TreeAdj(s) = \fbibrk{t : \exists \fbiangl{x,y} \in \ArcSet}$  for $x \in \Tree(s), y \in \Tree(t)$, and $s\neq t$. Finally, we can formulate two novel color-based criteria: $\SeedRelMinContr(s) = \min_{\forall t \in \TreeAdj(s)}\fbibrk{\TreeGradFeat(s,t)}$ and $\SeedRelMaxContr(s) = \max_{\forall t \in \TreeAdj(s)}\fbibrk{\TreeGradFeat(s,t)}$; which aims for the minimum and maximum contrast , respectively, amongst the seed's neighbors.

		Similarly to $\SeedRelSize$, considering only $\SeedRelMinContr$ or $\SeedRelMaxContr$ may not be sufficient for selecting relevant seeds. High contrast regions can indicate noises or a well-defined object border. Conversely, low contrast ones are often located within objects or nearby poorly-defined borders.  Thus, given that both size and contrast are important for describing the relevance of the seed, we propose two new criteria by combining the size-based and contrast-based estimations. The $\SeedRelMinSC(s) = \SeedRelSize(s)\cdot\SeedRelMinContr(s)$ favors large superpixels and prioritizes the minimum contrast amongst the superpixel neighbors, while $\SeedRelMaxSC(s) = \SeedRelSize(s)\cdot\SeedRelMaxContr(s)$ also favors large superpixels, but aims for the maximum contrast.

		Still, the relevance of a seed $s$ is strongly tied to which object is desired, making it irrelevant when it does not delineate the expected one. If an object is desired, then more superpixels should be within or near it, in order to promote its accurate delineation. In such case, the probable object location can assist in removing seeds that are not within (or nearby) the object of interest. First, analogously to $\TreeGradFeat(s,t)$, we define $\TreeGradObj(s,t) = \fblnorm{2}{\MeanObjFunc(\Tree(s)) - \MeanObjFunc(\Tree(t))}$, given $s,t \in \SeedSet$. Then, we propose an object-based criterion $\SeedRelObj(s) = \SeedRel_*(s)\cdot\max\fbibrk{\MeanObjFunc(\Tree(s)),\max_{t \in \TreeAdj(s)}\fbibrk{\TreeGradObj(s,t))}}$. By setting $\ObjFunc(p) = 1\; \forall p\in\VertexSet$ whenever a prior object information is abscent, one may see that $\SeedRelObj(s) = \SeedRel_*(s)$.

  \section{Experimental Results}
\label{sec:results}

	In this section, we detail the experimental framework for evaluating our proposal and discuss the obtained results. In Section~\ref{sec:results:setup}, the experimental setup is presented and, in Section~\ref{sec:results:ablation}, we perform an ablation study over all SICLE's steps. Finally, we compare our proposal against state-of-the-art methods both quantitatively and qualitatively (\fbie, Sections~\ref{sec:results:quantanal} and~\ref{sec:results:qualanal}).

	\subsection{Experimental Setup}
	\label{sec:results:setup}

		In order to evaluate the performance of each method for different domains, we selected four datasets whose object of interest is clearly defined (\textit{i.e.}, not ambiguous). The \textit{Extended Complex Saliency Scene Dataset}~(ECSSD)~\cite{Shi:2015:ECSSD} is a well-known salient object detection dataset composed of 1000 natural images with distinct and complex objects. Another natural image dataset, \textit{Insects}~\cite{Mansilla:2016:BirdsInsects}, has 130 images of insects whose thin legs impose a significant challenge in delineation. For evaluating objects with smooth borders, we opted for two different medical image datasets. The \textit{Liver}~\cite{Vargas:2019:ISF} dataset contains 40 CT slices of the human liver, whose grayscale borders are difficult to define. Finally, the \textit{Parasites}~\cite{Belem:2018:OISF} dataset has 72 colored images of helminth eggs in which an impurity may be attached to it. We randomly selected 70\% of the data for testing and 30\% for training. For generating the object saliency maps, we opted for an effective deep learning approach, named U$^2$-Net~\cite{Qin:2020:U2Net}, suitable for small datasets like Liver and Parasites. We trained such estimator using the training set and considering its default parameters.

		We selected several state-of-the-art superpixel methods based on their effectiveness and efficiency: (i) SLIC~\cite{Achanta:2012:SLIC}~\footnote{https://www.epfl.ch/labs/ivrl/research/slic-superpixels/}; (ii) LSC~\cite{Li:2015:LSC}~\footnote{https://jschenthu.weebly.com/projects.html}; (iii) SH~\cite{Wei:2018:SH}~\footnote{https://github.com/semiquark1/boruvka-superpixel}; (iv) ERS~\cite{Liu:2011:ERS}~\footnote{https://github.com/mingyuliutw/EntropyRateSuperpixel}; (v) OISF-OSMOX~\cite{Belem:2020:OISF}~\footnote{https://github.com/LIDS-UNICAMP/OISF}; and (vi) SEAL-ERS~\cite{Tu:2018:SEAL}~\footnote{https://github.com/wctu/SEAL}. For all methods, the recommended parameter configuration was set. Our code will be publicly available~\footnote{https://github.com/LIDS-UNICAMP/SICLE} after publishing. Given that it is burdensome to evaluate all possible SICLE configurations, we elected the following one as starting point for our analysis based on previous works. We sampled using GRID with $\Ninit = 8000$ seeds and considered $\ArcFuncDyn$ as in~\cite{Belem:2020:DISF}, and used the $\SeedRelObj$, considering $\SeedRelMinSC$, as presented in~\cite{Belem:2021:ODISF}. Finally, similarly to most seed-based methods~\cite{Achanta:2012:SLIC,Li:2015:LSC,Vargas:2019:ISF}, we set $\Omega = 10$.

		Due to the high correlation between several superpixel segmentation metrics~\cite{Stutz:2018:Superpixels}, we chose the two most popular. The \textit{Boundary Recall}~(BR)~\cite{Stutz:2018:Superpixels} calculates the percentage of the object boundaries correctly overlapped by a superpixel border given a certain tolerance radius. The \textit{Under-Segmentation Error}~(UE)~\cite{Neubert:2012:Superpixel} measures the error resultant from superpixels overlapping multiple ground-truths. We aim for higher values of BR and lower values of UE. For all experiments, we set $\Nfinal \in \fbisqbrk{25,750}$.

	\subsection{Ablation Study}
	\label{sec:results:ablation}

\begin{figure}[ht!]
			\centering
			\includegraphics[page = 1, width = \textwidth, trim = {6.5 0 8 0}, clip]%
			{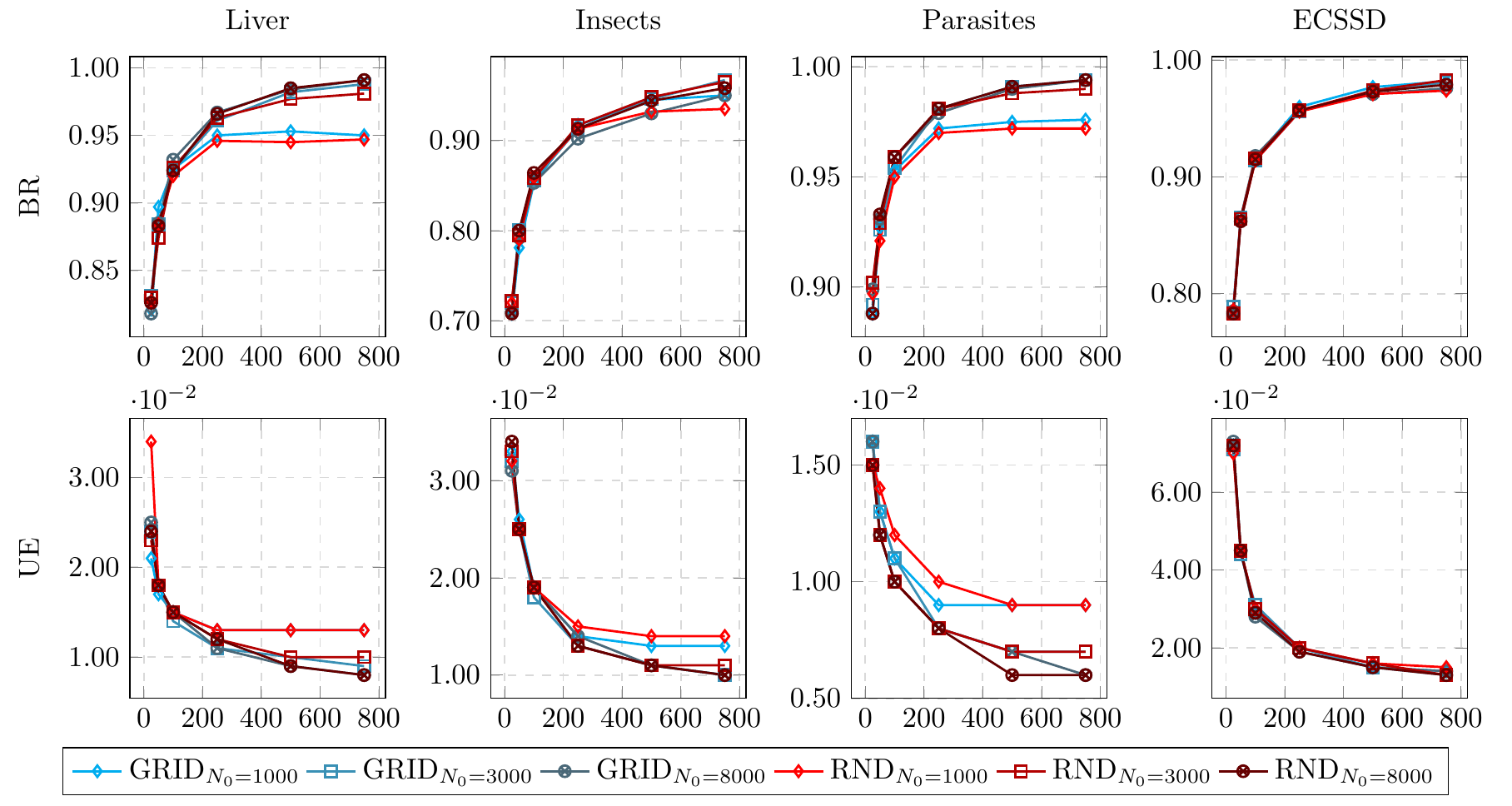}
			\caption{Results of SICLE variants with varying oversampling strategy and $\Ninit$.}
			\label{fig:res:exper1}
		\end{figure}

		We analyze the impacts on the oversampling task by altering the strategy (\fbie, GRID or RND) and $\Ninit$, as Figure~\ref{fig:res:exper1} shows. It is possible to see that both strategies achieved a similar performance, irrespective of $\Ninit$, leading to conclude that pattern has no substantial influence in SICLE. On the other hand, we may also affirm that there are multiple ways to obtain a relevant seed set, given that different strategies achieved equivalent performance. That is, such result support the idea of relevance redundancy amongst seeds.

		In terms of $\Ninit$, it is possible to see that $1000$ seeds are sufficient for accurate delineation in the ECSSD dataset but insufficient for the remaining ones. We argue that such event is due to the images' size: the bigger the images, the more seeds are required in oversampling. However, $8000$ is unnecessary for all datasets considered, especially given that $5000$ achieves on par performance with fewer seeds. Thus, based on efficiency and simplicity, we opt for randomly oversampling $3000$ seeds.

\begin{figure}[ht!]
			\centering
			\includegraphics[page = 2, width = \textwidth, trim = {6.5 0 8 0}, clip]%
			{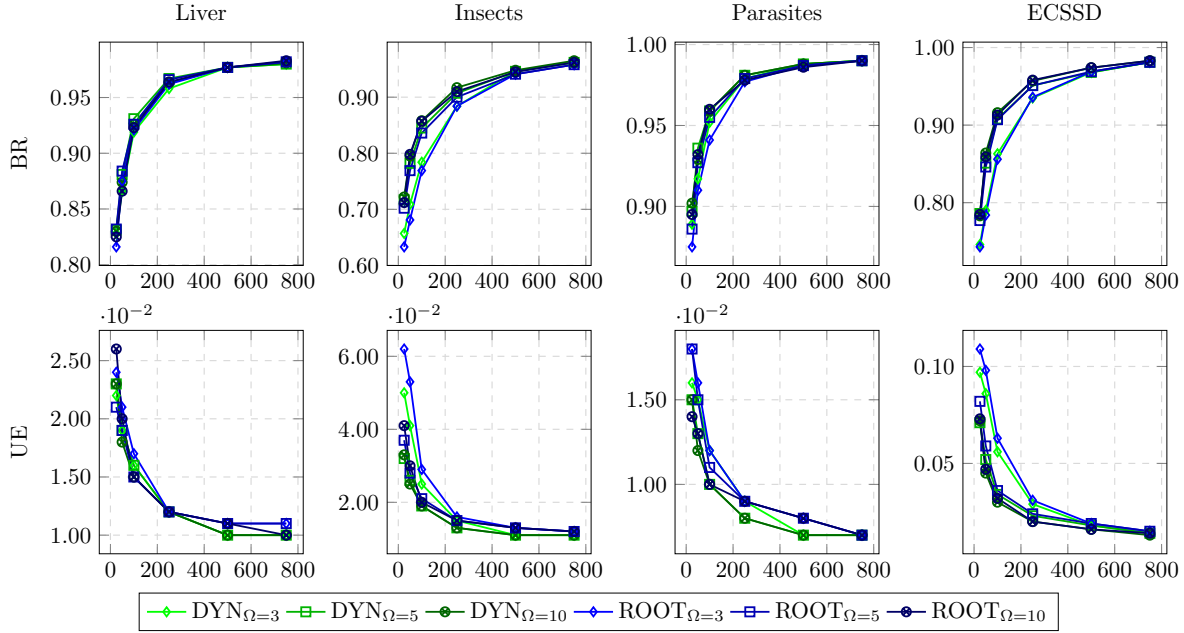}
			\caption{Results of SICLE variants with varying arc-cost functions and $\Omega$.}
			\label{fig:res:exper2}
		\end{figure}

		Figure~\ref{fig:res:exper2} shows the results for varying arc-cost functions (\fbie, DYN and ROOT) and limiting $\Omega \in \fbibrk{3,5,10}$. The equal performance between estimations can be explained by the possible similarity between the seed's and tree's features. Interestingly, $\Omega=3$ is sufficient for homogeneous objects (\fbie, liver and parasite egg) but insufficient for more complex ones. We argue that the number of iterations for seed relevance manifestation is proportional to the object's complexity. However, when comparing $\Omega=5$ and $\Omega=10$ in all datasets, there are no significant improvements, indicating a possible upper bound (\fbie, performing unnecessary iterations). Therefore, we limited SICLE to perform, at most, five iterations using the ROOT function, which allows future improvements~\cite{Goncalves:2019:cudaIFT,Falcao:2004:DIFT}.

\begin{figure}[ht!]
			\centering
			\includegraphics[page = 3, width = \textwidth, trim = {6.5 0 8 0}, clip]%
			{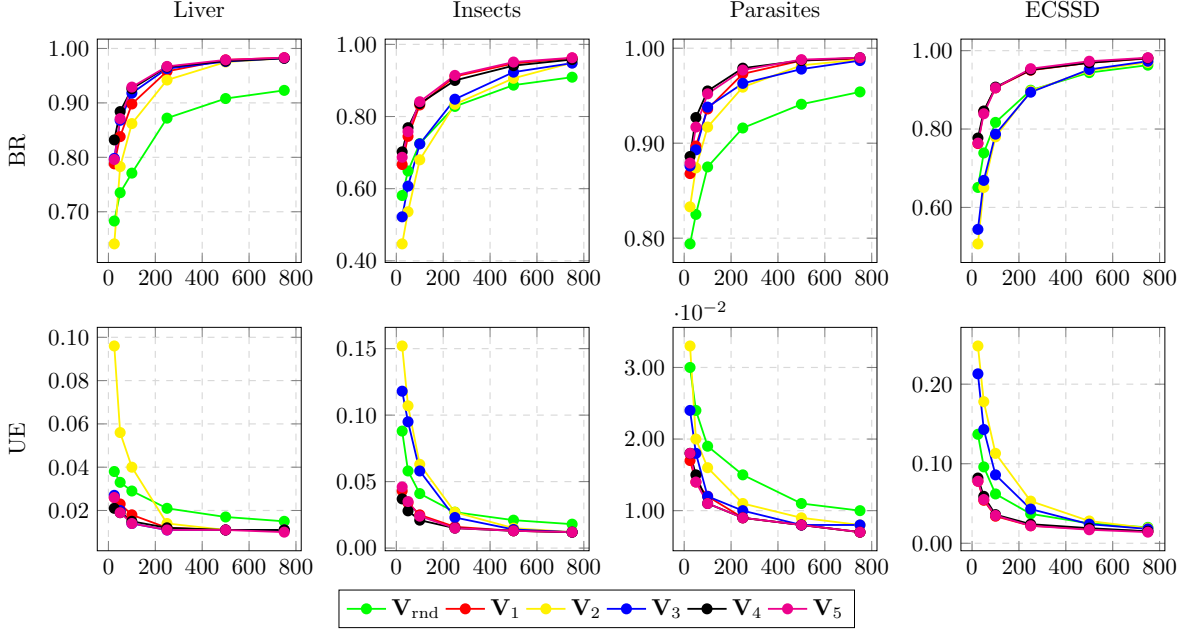}
			\caption{Results of SICLE variants with varying seed removal criteria.}
			\label{fig:res:exper3}
		\end{figure}

		We have also evaluated our proposed seed removal criteria $\fbibrk{\SeedRel_i}^{5}_{i=1}$ subjected to $\SeedRelObj$ alongside a random selection criterion $\SeedRelRnd$ as baseline (Figure~\ref{fig:res:exper3}). Unsurprisingly, the random criterion is the worst approach for all datasets. However, it is interesting that, while contrast-only criteria performs fairly for datasets in which contrast is crucial (\fbie, Liver and Parasites), size-only criterion achieves top performance in all datasets. We argue that $\SeedRelObj$ manages to overcome their possible biases by constraining the seed selection nearby (or within) the desired object. When combining both contrast and size estimations, we can see a slight improvement for Liver and Parasites, and no performance degradation for the remaining ones. Furthermore, if no map is provided, such combination can overcome their particular biases~\cite{Belem:2020:DISF}. Thus, we opt for maintaining the $\SeedRelMinSC$ subjected to $\SeedRelObj$.

\begin{figure}[ht!]
			\centering
			\includegraphics[page = 4, width = \textwidth, trim = {6.5 0 8 0}, clip]%
			{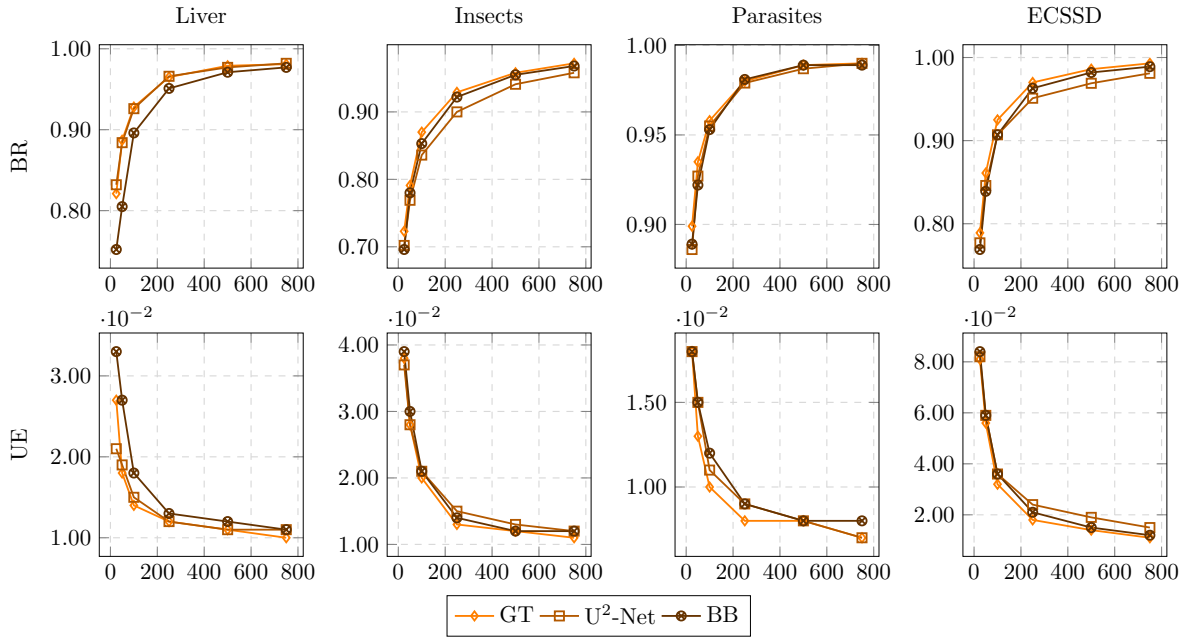}
			\caption{Results of SICLE variants with different object saliency maps.}
			\label{fig:res:exper4}
		\end{figure}

		Finally, as Figure~\ref{fig:res:exper4} shows, we assessed SICLE considering maps of different qualities for analyzing the influence of the map's borders in delineation. Aside from those generated by the U$^2$-Net, used throughout this paper, we considered the ground-truth (GT) and the object's minimum bounding box (BB) for emulating the generation of maps with ideal and rough object borders, respectively. We emphasize that the GT maps are only used in this particular analysis. It is important to note that, differently from other object-based strategies~\cite{Belem:2018:OISF,Belem:2020:OISF}, the delineation performance of SICLE is not impacted by the quality of the map's border. As consequence, and setting aside that generating ideal maps is yet an unfeasible task, we claim that such property is positive since it disregards the need to improve the object saliency map whenever it is not the problem's main goal.

	\subsection{Quantitative Analysis}
	\label{sec:results:quantanal}

\begin{figure}[t!]
		\centering
		\includegraphics[page = 5, width = \textwidth, trim = {6.5 0 8 0}, clip]%
		{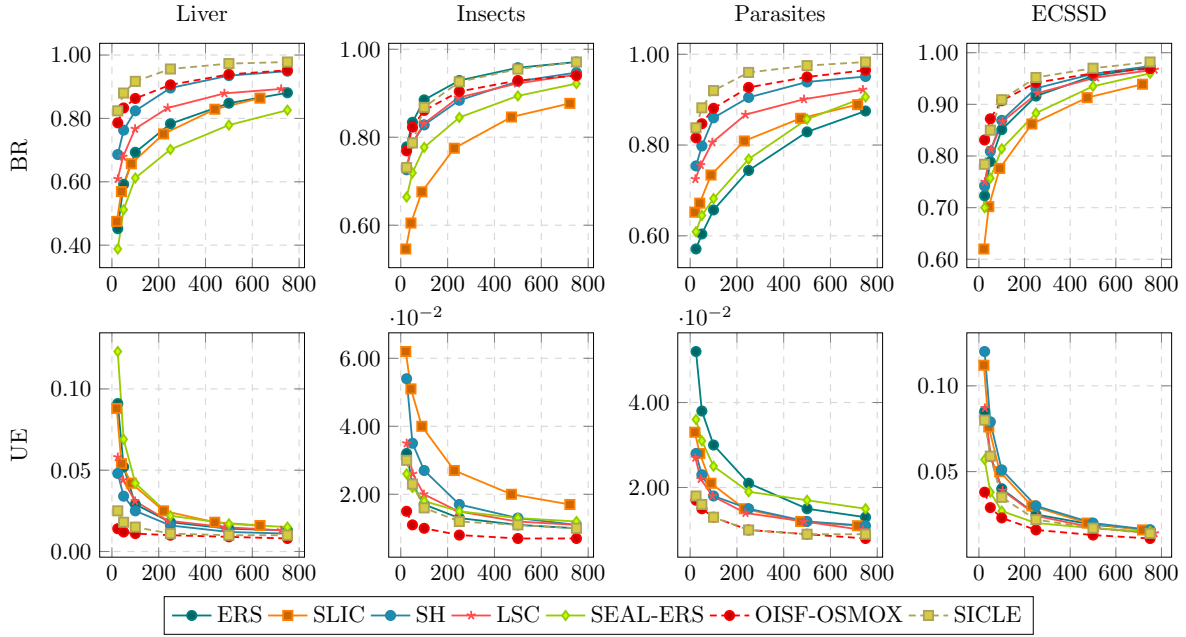}
		\caption{Results of SICLE against several state-of-the-art methods.}
		\label{fig:res:exper5}
	\end{figure}

	We compared our best SICLE variant against several state-of-the-art methods, as illustrated in Figure~\ref{fig:res:exper5}. It is possible to see that SICLE is consistently accurate, for both metrics, in all datasets. That is, considering different objects with distinct features, our approach's strategy of oversampling and object-based seed removal is proven more effective than the typical approach of seed recomputation seen in SLIC, LSC and OISF. When analyzing the UE, the results support the findings of~\cite{Belem:2021:ODISF,Belem:2019:OSMOX} claiming that object-based strategies offer a better superpixel leaking prevention.

	Table~\ref{tab:res:speed} shows the computational time required for each method on a 64bit Intel (R) Core(TM) i5-5200U PC with CPU speed of 2.20Ghz. In terms of time complexity, SICLE is $\Ocomplex{|\VertexSet|\log|\VertexSet|}$ bounded by the IFT executions. Although SLIC is $\Ocomplex{|\VertexSet|}$, we can see that, in practice, SICLE tends to be faster, since it is limited by, at most, 5 iterations. One could argue in favor of SH, which is $\Ocomplex{|\VertexSet|}$, given its property of computing a multiscale segmentation in a single execution. However, SICLE has such property if one sets the desired quantity of superpixels (\fbie, seeds) at each iteration. Lastly, SICLE is a more efficient object-based solution than OISF.

	\begin{table}[t!]
		\centering
		\begin{tabular}{c|c|c|c|c}
	$\Nfinal$ & SLIC & SH & OISF-OSMOX & SICLE\\\hline
	25 	& \textbf{0.472$\pm$0.026} & 0.916$\pm$0.035 & 3.091$\pm$0.628 & 0.541$\pm$0.067 \\
	100 & 0.473$\pm$0.025 & 0.915$\pm$0.032 & 1.757$\pm$0.343 & \textbf{0.445$\pm$0.060} \\
	750 & 0.477$\pm$0.025 & 0.916$\pm$0.032 & 1.182$\pm$0.229 & \textbf{0.361$\pm$0.053} \\
\end{tabular}

		\caption{Average execution time for each method in the ECSSD dataset. The best performance for each quantity is depicted in bold.}
		\label{tab:res:speed}
	\end{table}

	\subsection{Qualitative Analysis}
	\label{sec:results:qualanal}
	\begin{sidewaysfigure}
		\centering
		\includegraphics[width = 0.98\textheight]{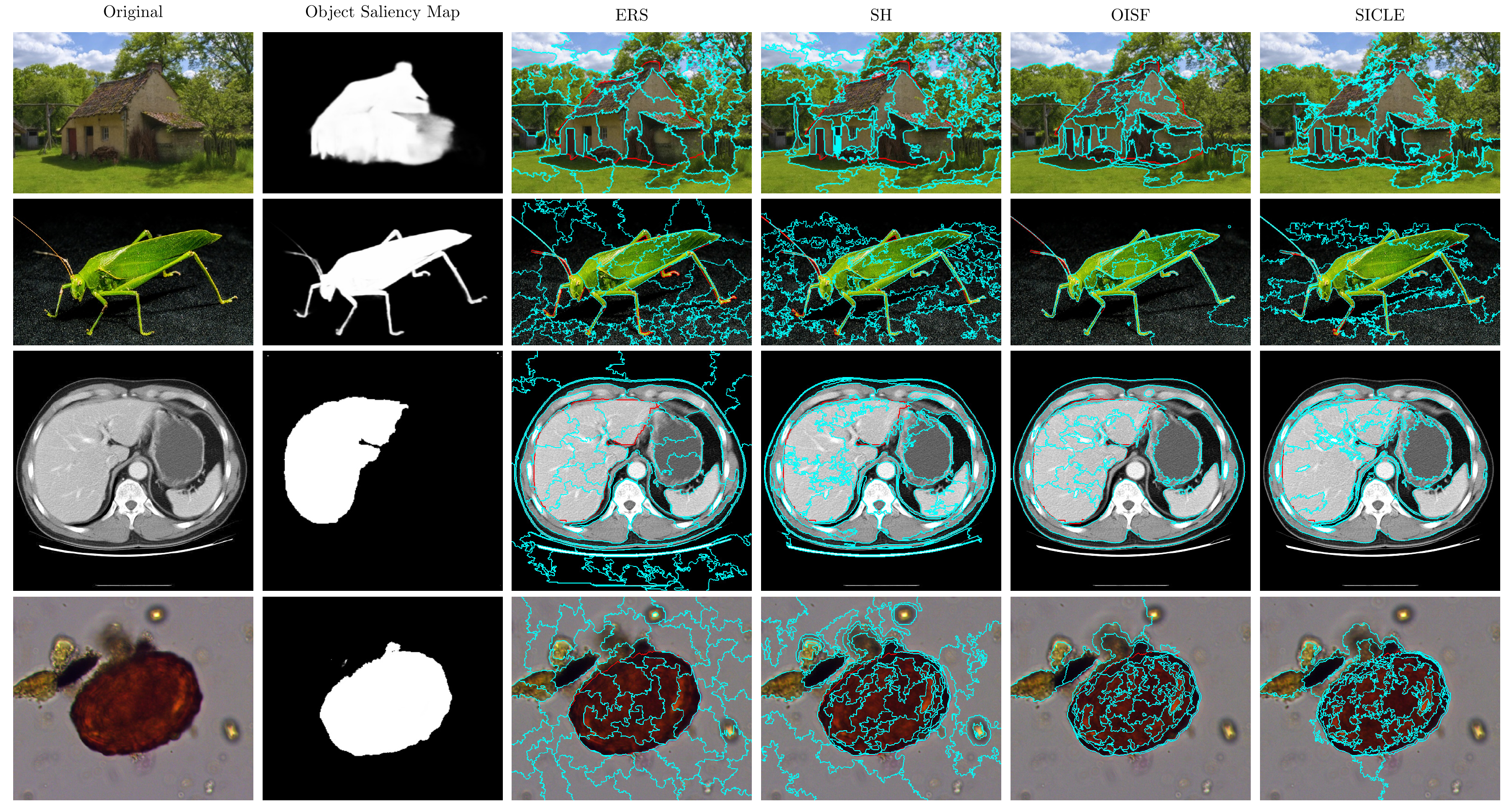}
		\caption{Qualitative comparison between SICLE and several state-of-the-art methods. The desired number of superpixels was $\Nfinal=50$. The red lines indicate the object boundary, whereas the cyan ones, superpixel borders.}
		\label{fig:res:qualanal}
	\end{sidewaysfigure}

		We compare some segmentation results between SICLE and other state-of-the-art methods in Figure~\ref{fig:res:qualanal}. The ERS method achieves effective object delineation only for the insect image, whereas for the other images, it performs fairly. Conversely, we can see that SH present better and, more importantly, more consistent results for all images. However, the lack of object information compromises the object delineation of the desired object in several parts. For example, both object-based methods increased the superpixel resolution within (or nearby) the object and improved its delineation. Still, for OISF, the incorrect estimations in the map assisted in deteriorating the object delineation in several parts. Finally, it is possible to see that SICLE achieves effective delineation for all objects irrespective of saliency errors.

  \section{Conclusion and Future Work}
\label{sec:conclusion}

	In this work, we propose a novel object-based superpixel segmentation framework named \textit{Superpixels through Iterative CLEarcutting}~(SICLE), which is a generalization of two state-of-the-art methods~\cite{Belem:2020:DISF,Belem:2021:ODISF}. It starts off from oversampling and, through several iterations, generates superpixels from the seed set and remove a portion of irrelevant seeds for preserving the accurate object delineation from the previous iteration. We show that SICLE is consistently effective by evaluating its performance against several state-of-the-art methods in different datasets. Moreover, results show that, in most practical cases, SICLE can be faster than SLIC. For future endeavors, we intent to extend SICLE for 3D and video segmentation, while exploring its capability of being an interactive segmentation tool.

\section*{Acknowledgment}
 
 
 The authors thank the Conselho Nacional de Desenvolvimento Cient{\'i}fico e Tecnol{\'o}gico -- CNPq -- (Universal 407242/2021-0, PQ 303808/2018-7, 310075/2019-0), the Funda\c{c}{\~a}o de Amparo a Pesquisa do Estado de Minas Gerais -- FAPEMIG -- (PPM-00006-18), the Funda\c{c}{\~a}o de Amparo a Pesquisa do Estado de S{\~a}o Paulo -- FAPESP -- (2014/12236-1) and the Coordena{\c c}{\~a}o de Aperfei{\c c}oamento de Pessoal de N{\'i}vel Superior -- CAPES -- Finance code 001 (COFECUB 88887.191730/2018-00) for the financial support.

  \bibliographystyle{abbrv}
  \bibliography{refs/superpixels.bib,refs/others.bib}

\begin{thebibliography}{10}

\bibitem{Achanta:2012:SLIC}
R.~Achanta, A.~Shaji, K.~Smith, A.~Lucchi, P.~Fua, and S.~S{\"u}sstrunk.
\newblock {SLIC} superpixels compared to state-of-the-art superpixel methods.
\newblock {\em Transactions on Pattern Analysis and Machine Intelligence},
  34(11):2274--2282, 2012.

\bibitem{Achanta:2017:SNIC}
R.~Achanta and S.~S{\"u}sstrunk.
\newblock Superpixels and polygons using simple non-iterative clustering.
\newblock In {\em 30th Conference on Computer Vision and Pattern Recognition
  (CVPR)}, pages 4651--4660, 2017.

\bibitem{Awaisu:2019:DeepFLIC}
M.~Awaisu, L.~Li, J.~Peng, and J.~Zhang.
\newblock Fast superpixel segmentation with deep features.
\newblock In {\em 36th Computer Graphics International Conference (CGI)}, pages
  410--416, 2019.

\bibitem{Ban:2018:GMM}
Z.~Ban, J.~Liu, and L.~Cao.
\newblock Superpixel segmentation using gaussian mixture model.
\newblock {\em Transactions on Image Processing}, 27(8):4105--4117, 2018.

\bibitem{Belem:2021:ODISF}
F.~Bel{\'e}m, J.~Cousty, B.~Perret, S.~Guimar{\~a}es, and A.~Falc{\~a}o.
\newblock Towards a simple and efficient object-based superpixel delineation
  framework.
\newblock In {\em 34th Conference on Graphics, Patterns and Images (SIBGRAPI)},
  pages 346--353, 2021.

\bibitem{Belem:2018:OISF}
F.~Bel{\'e}m, S.~Guimar{\~a}es, and A.~Falc{\~a}o.
\newblock Superpixel segmentation by object-based iterative spanning forest.
\newblock In {\em 23rd Iberoamerican Congress on Pattern Recognition}, pages
  334--341, 2018.

\bibitem{Belem:2020:OISF}
F.~Bel{\'e}m, S.~Guimar{\~a}es, and A.~Falc{\~a}o.
\newblock Superpixel generation by the iterative spanning forest using object
  information.
\newblock In {\em 33rd Conference on Graphics, Patterns and Images (SIBGRAPI)},
  pages 22--28, 2020.
\newblock Workshop of Thesis and Dissertations.

\bibitem{Belem:2020:DISF}
F.~Bel{\'e}m, S.~Guimar{\~a}es, and A.~Falc{\~a}o.
\newblock Superpixel segmentation using dynamic and iterative spanning forest.
\newblock {\em Signal Processing Letters}, 27:1440--1444, 2020.

\bibitem{Belem:2019:OSMOX}
F.~Bel{\'e}m, L.~Melo, S.~Guimar{\~a}es, and A.~Falc{\~a}o.
\newblock The importance of object-based seed sampling for superpixel
  segmentation.
\newblock In {\em 32nd Conference on Graphics, Patterns and Images (SIBGRAPI)},
  pages 108--115, 2019.

\bibitem{Bragantini:2018:DynIFT}
J.~Bragantini, S.~Martins, C.~Castelo-Fernandez, and A.~Falc{\~a}o.
\newblock Graph-based image segmentation using dynamic trees.
\newblock In {\em 23rd Iberoamerican Congress on Pattern Recognition}, pages
  470--478, 2018.

\bibitem{Castelo:2019:ClassISF}
C.~Castelo-Fernandez and A.~Falc{\~a}o.
\newblock Learning visual dictionaries from class-specific superpixel
  segmentation.
\newblock In {\em 18th Conference on Computer Analysis of Images and Patterns
  (CAIP)}, pages 171--182, 2019.

\bibitem{Dhore:2021:Xray}
S.~Dhore and D.~Abin.
\newblock Chest x-ray segmentation using watershed and super pixel segmentation
  technique.
\newblock In {\em 4th International Conference on Communication information and
  Computing Technology (ICCICT)}, pages 1--4, 2021.

\bibitem{Falcao:2004:DIFT}
A.~Falc{\~a}o and F.~Bergo.
\newblock Interactive volume segmentation with differential image foresting
  transforms.
\newblock {\em Transactions on Medical Imaging}, 23(9):1100--1108, 2004.

\bibitem{Falcao:2004:IFT}
A.~Falc{\~a}o, J.~Stolfi, and R.~Lotufo.
\newblock The image foresting transform: Theory, algorithms, and applications.
\newblock {\em Transactions on Pattern Analysis and Machine Intelligence},
  26(1):19--29, 2004.

\bibitem{Galvao:2018:RISF}
F.~Galv{\~a}o, A.~Falc{\~a}o, and A.~Chowdhury.
\newblock {RISF:} recursive iterative spanning forest for superpixel
  segmentation.
\newblock In {\em 31st Conference on Graphics, Patterns and Images (SIBGRAPI)},
  pages 408--415, 2018.

\bibitem{Goncalves:2019:cudaIFT}
H.~Gon{\c{c}}alves, G.~Vasconcelos, P.~Rangel, M.~Carvalho, N.~Archilha, and
  T.~Spina.
\newblock cudaift: 180x faster image foresting transform for waterpixel
  estimation using cuda.
\newblock In {\em 14th International Conference on Computer Graphics Theory and
  Applications (VISIGRAPP)}, pages 395--404, 2019.

\bibitem{Jampani:2018:SSN}
V.~Jampani, D.~Sun, M.~Liu, M.~Yang, and J.~Kautz.
\newblock Superpixel sampling networks.
\newblock In {\em 18th European Conference on Computer Vision (ECCV)}, pages
  352--368, 2018.

\bibitem{Li:2015:LSC}
Z.~Li and J.~Chen.
\newblock Superpixel segmentation using linear spectral clustering.
\newblock In {\em 28th Conference on Computer Vision and Pattern Recognition
  (CVPR)}, pages 1356--1363, 2015.

\bibitem{Liu:2019:LungSegmentation}
C.~Liu, R.~Zhao, and M.~Pang.
\newblock A fully automatic segmentation algorithm for ct lung images based on
  random forest.
\newblock {\em Medical Physics}, 2019.

\bibitem{Liu:2011:ERS}
M.~Liu, O.~Tuzel, S.~Ramalingam, and R.~Chellappa.
\newblock Entropy rate superpixel segmentation.
\newblock In {\em 24th Conference on Computer Vision and Pattern Recognition
  (CVPR)}, pages 2097--2104, 2011.

\bibitem{Liu:2018:IMSLIC}
Y.~Liu, M.~Yu, B.~Li, and Y.~He.
\newblock Intrinsic manifold {SLIC}: A simple and efficient method for
  computing content-sensitive superpixels.
\newblock {\em Transactions on Pattern Analysis and Machine Intelligence},
  40(3):653--666, 2018.

\bibitem{Mansilla:2016:BirdsInsects}
L.~Mansilla and P.~Miranda.
\newblock Oriented image foresting transform segmentation: Connectivity
  constraints with adjustable width.
\newblock In {\em 29th Conference on Graphics, Patterns and Images (SIBGRAPI)},
  pages 289--296, 2016.

\bibitem{Martins:2019:SymmISF}
S.~Martins, G.~Ruppert, F.~Reis, C.~Yasuda, and A.~Falc{\~a}o.
\newblock A supervoxel-based approach for unsupervised abnormal asymmetry
  detection in {MR} images of the brain.
\newblock In {\em 16th International Symposium on Biomedical Imaging (ISBI)},
  pages 882--885, 2019.

\bibitem{Neubert:2012:Superpixel}
P.~Neubert and P.~Protzel.
\newblock Superpixel benchmark and comparison.
\newblock In {\em Forum Bildverarbeitung}, volume~6, pages 1--12, 2012.

\bibitem{Peng:2021:HERS}
H.~Peng, A.~Aviles-Rivero, and C.~Schonlieb.
\newblock Hers superpixels: Deep affinity learning for hierarchical entropy
  rate segmentation.
\newblock {\em arXiv preprint}, 2021.

\bibitem{Qin:2020:U2Net}
X.~Qin, Z.~Zhang, C.~Huang, M.~Dehghan, O.~Zaiane, and M.~Jagersand.
\newblock U2-net: Going deeper with nested u-structure for salient object
  detection.
\newblock {\em Pattern Recognition}, 106:107404, 2020.

\bibitem{Shen:2016:DBSCAN}
J.~Shen, X.~Hao, Z.~Liang, Y.~Liu, W.~Wang, and L.~Shao.
\newblock Real-time superpixel segmentation by {DBSCAN} clustering algorithm.
\newblock {\em Transactions on Image Processing}, 25(12):5933--5942, 2016.

\bibitem{Shi:2015:ECSSD}
J.~Shi, Q.~Yan, L.~Xu, and J.~Jia.
\newblock Hierarchical image saliency detection on extended cssd.
\newblock {\em Transactions on Pattern Analysis and Machine Intelligence},
  38(4):717--729, 2015.

\bibitem{Sousa:2019:ALTIS}
A.~Sousa, S.~Martins, A.~Falc{\~a}o, F.~Reis, E.~Bagatin, and K.~Irion.
\newblock Altis: A fast and automatic lung and trachea ct-image segmentation
  method.
\newblock {\em Medical Physics}, 46(11):4970--4982, 2019.

\bibitem{Stutz:2018:Superpixels}
D.~Stutz, A.~Hermans, and B.~Leibe.
\newblock Superpixels: An evaluation of the state-of-the-art.
\newblock {\em Computer Vision and Image Understanding}, 166:1--27, 2018.

\bibitem{Suzuki:2020:RIM}
T.~Suzuki.
\newblock Superpixel segmentation via convolutional neural networks with
  regularized information maximization.
\newblock In {\em 45th International Conference on Acoustics, Speech and Signal
  Processing (ICASSP)}, pages 2573--2577, 2020.

\bibitem{Tu:2018:SEAL}
W.~Tu, M.~Liu, V.~Jampani, D.~Sun, S.~Chien, M.~Yang, and J.~Kautz.
\newblock Learning superpixels with segmentation-aware affinity loss.
\newblock In {\em 31st Conference on Computer Vision and Pattern Recognition
  (CVPR)}, pages 568--576, 2018.

\bibitem{Vargas:2019:ISF}
J.~Vargas-Mu{\~n}oz, A.~Chowdhury, E.~Alexandre, F.~Galv{\~a}o, P.~Miranda, and
  A.~Falc{\~a}o.
\newblock An iterative spanning forest framework for superpixel segmentation.
\newblock {\em Transactions on Image Processing}, 28(7):3477--3489, 2019.

\bibitem{Wei:2018:SH}
X.~Wei, Q.~Yang, Y.~Gong, N.~Ahuja, and M.~Yang.
\newblock Superpixel hierarchy.
\newblock {\em Transactions on Image Processing}, 27(10):4838--4849, 2018.

\bibitem{Wu:2021:TASP}
J.~Wu, C.~Liu, and B.~Li.
\newblock Texture-aware and structure-preserving superpixel segmentation.
\newblock {\em Computers and Graphics}, 94:152--163, 2021.

\bibitem{Xiao:2018:CAS}
X.~Xiao, Y.~Zhou, and Y.~Gong.
\newblock Content-adaptive superpixel segmentation.
\newblock {\em Transactions on Image Processing}, 27(6):2883--2896, 2018.

\bibitem{Xu:2014:SS}
L.~Xu, L.~Zeng, and Z.~Wang.
\newblock Saliency-based superpixels.
\newblock {\em Signal, Image and Video Processing}, 8(1):181--190, 2014.

\bibitem{Yang:2020:FCN}
F.~Yang, Q.~Sun, H.~Jin, and Z.~Zhou.
\newblock Superpixel segmentation with fully convolutional networks.
\newblock In {\em 33rd Conference on Computer Vision and Pattern Recognition
  (CVPR)}, 2020.

\bibitem{Yi:2022:Semantic}
S.~Yi, H.~Ma, X.~Wang, T.~Hu, X.~Li, and Y.~Wang.
\newblock Weakly-supervised semantic segmentation with superpixel guided local
  and global consistency.
\newblock {\em Pattern Recognition}, 124:108504, 2022.

\bibitem{Yu:2019:Pedestrian}
Y.~Yu, Y.~Makihara, and Y.~Yagi.
\newblock Pedestrian segmentation based on a spatio-temporally consistent
  graph-cut with optimal transport.
\newblock {\em Transactions on Computer Vision and Applications}, 11(1):10,
  2019.

\bibitem{Yu:2021:EASS}
Y.~Yu, Y.~Yang, and K.~Liu.
\newblock Edge-aware superpixel segmentation with unsupervised convolutional
  neural networks.
\newblock In {\em 28th International Conference on Image Processing (ICIP)},
  pages 1504--1508, 2021.

\bibitem{Zhang:2018:Plant}
S.~Zhang, H.~Wang, W.~Huang, and Z.~You.
\newblock Plant diseased leaf segmentation and recognition by fusion of
  superpixel, {K}-means and {PHOG}.
\newblock {\em Optik}, 157:866--872, 2018.

\bibitem{Zhou:2019:Breast}
J.~Zhou, J.~Ruan, C.~Wu, G.~Ye, Z.~Zhu, J.~Yue, and Y.~Zhang.
\newblock Superpixel segmentation of breast cancer pathology images based on
  features extracted from the autoencoder.
\newblock In {\em 11th International Conference on Communication Software and
  Networks (ICCSN)}, pages 366--370, 2019.

\end{thebibliography}

\end{document}